\documentclass[10pt, a4paper]{article}

\usepackage{lrec-coling2024} % this is the new style
\usepackage{multirow}
\usepackage{float}
\usepackage{booktabs}

\usepackage{graphicx}
\newcommand{\DEVELOPMENT}{1} % 1= show comments, 0=no comments
\usepackage{ifthen}
\ifthenelse{\DEVELOPMENT = 1}{
  \newcommand{\tz}[1]{\textcolor{red}{\textbf{TZ:} #1}}
}{
\newcommand{\tz}[1]{}
}
\ifthenelse{\DEVELOPMENT = 1}{
  \newcommand{\ka}[1]{\textcolor{blue}{\textbf{KA:} #1}}
}{
\newcommand{\ka}[1]{}
}

\title{Comprehensive Study on German Language Models for Clinical and Biomedical Text Understanding}

\name{Ahmad Idrissi-Yaghir$^{1,2\ast}$, Amin Dada$^{3\ast}$, Henning Schäfer$^{1,4\ast}$,\\ {\bf \large Kamyar Arzideh}$^{3}$, 
{\bf \large Giulia Baldini}$^{3,11}$, {\bf \large Jan Trienes}$^{3}$, {\bf \large Max Hasin}$^{3}$, \\{\bf \large Jeanette Bewersdorff}$^{5}$, {\bf \large Cynthia S. Schmidt$^{3,4}$}, {\bf \large Marie Bauer}$^{3}$, \\{\bf \large Kaleb E. Smith$^{6}$}, {\bf \large Jiang Bian$^{7,8}$}, {\bf \large Yonghui Wu$^{7,8}$}, {\bf \large Jörg Schlötterer$^{9,10}$}, \\{\bf \large Torsten Zesch$^{5}$}, {\bf \large Peter A. Horn$^{4}$}, {\bf \large Christin Seifert$^{9}$},  \\{\bf \large Felix Nensa$^{3,11}$}, {\bf \large Jens Kleesiek$^{3,12,13,14}$},{\bf \large Christoph M. Friedrich$^{1,2}$}}

\address{\small $^{1}$ Department of Computer Science, University of Applied Sciences and Arts Dortmund, Dortmund, Germany  \\
\small $^{2}$ Institute for Medical Informatics, Biometry and Epidemiology, University Hospital Essen, Essen, Germany \\
\small $^{3}$ Institute for AI in Medicine (IKIM), University Hospital Essen (AöR), Essen, Germany \\
\small $^{4}$Institute for Transfusion Medicine, University Medicine Essen, Essen, Germany \\
\small $^{5}$Computational Linguistics, CATALPA FernUniversität in Hagen, Germany \\
\small $^{6}$NVIDIA, Santa Clara, CA, USA \\
\small $^{7}$Department of Health Outcomes and Biomedical Informatics, College of Medicine,\\ \small University of Florida, Gainesville, FL, USA \\
\small$^{8}$Cancer Informatics and eHealth core, University of Florida Health Cancer Center, \\ \small University of Florida, Gainesville, FL, USA \\ 
\small$^{9}$University of Marburg, Marburg, Germany \\
\small$^{10}$University of Mannheim, Mannheim, Germany \\
\small$^{11}$University Hospital Essen, Institute of Interventional and Diagnostic Radiology and Neuroradiology, \\ \small Essen, Germany \\
\small$^{12}$Cancer Research Center Cologne Essen (CCCE), West German Cancer Center Essen,\\ \small University Hospital Essen (AöR), Essen, Germany \\
\small $^{13}$German Cancer Consortium (DKTK, Partner site Essen), Heidelberg, Germany \\
\small $^{14}$Department of Physics, TU Dortmund, Dortmund, Germany \\
\small \{ahmad.idrissi-yaghir, christoph.friedrich\}@fh-dortmund.de \\
\small \{amin.dada, henning.schaefer, jens.kleesiek\}@uk-essen.de\\}

\abstract{
Recent advances in natural language processing (NLP) can be largely attributed to the advent of pre-trained language models such as BERT and RoBERTa. While these models demonstrate remarkable performance on general datasets, they can struggle in specialized domains such as medicine, where unique domain-specific terminologies, domain-specific abbreviations, and varying document structures are common. This paper explores strategies for adapting these models to domain-specific requirements, primarily through continuous pre-training on domain-specific data. We pre-trained several German medical language models on 2.4B tokens derived from translated public English medical data and 3B tokens of German clinical data. The resulting models were evaluated on various German downstream tasks, including named entity recognition (NER), multi-label classification, and extractive question answering. Our results suggest that models augmented by clinical and translation-based pre-training typically outperform general domain models in medical contexts. We conclude that continuous pre-training has demonstrated the ability to match or even exceed the performance of clinical models trained from scratch. Furthermore, pre-training on clinical data or leveraging translated texts have proven to be reliable methods for domain adaptation in medical NLP tasks.
 \\ \newline \Keywords{German-centric NLP, Clinical Language Models, Domain Adaptation} 
}
\begin{document}

\maketitleabstract

\section{Introduction}
\renewcommand{\thefootnote}{\fnsymbol{footnote}}
\footnote[0]{$^{\ast}$ These authors contributed equally to this work}
\renewcommand{\thefootnote}{\arabic{footnote}}
In recent years, pre-trained language models (PLMs) such as BERT \cite{DevlinCLT19} and RoBERTa \cite{Liu19-roberta} have become crucial in the field of natural language processing (NLP). These models have significantly enhanced the performance of a wide range of tasks, including document and token classification, as well as machine translation and text summarization. The success of these models is primarily attributed to their transformer-based architecture \cite{VaswaniSPUJGKP17} and their training on large amounts of unlabeled data. However, since these models are often trained on general data sources such as Wikipedia, news articles, and books, their effectiveness may be limited in specific domains, such as medicine or finance, which have distinct terminologies and writing styles. To achieve better results in these specialized fields, it is essential to use language models that are tailored to these specific domains.
Building on this concept, language models can be adapted to specialized domains through two methods. The first approach involves training models from scratch on unlabeled data from the desired domain. This method ensures that the model is grounded in the unique characteristics of the target domain from the beginning. The second approach relies on continuous pre-training. Instead of starting from scratch, existing general-purpose pre-trained models can be used and refined through further pre-training on the domain-specific unlabeled data \cite{DBLP:conf/acl/GururanganMSLBD20}. This allows a transition that shifts the focal point of the model from a broad scope to one specific to the particularities of the target domain.

Particularly in the medical domain, such specialized models can potentially improve the practice of medicine by providing accurate and relevant insights from vast amounts of textual data. These specialized models are highly valuable given the complex nature of medical terminology and the critical importance of accurate information in healthcare. For example, they can help analyze patient records, extract critical information from medical literature, and facilitate real-time clinical decision-making by understanding patient queries or medical notes.

However, building these specialized medical models presents unique challenges. Medical data is characterized not only by specialized terminology but also by the sensitive nature of the information. Patient confidentiality and other ethical considerations are paramount, which can complicate the acquisition of large datasets for training. 

Particular focus has been set on German medical data because of its data sparsity when compared to English datasets \cite{otherenglish, biobertpt}. While resources for languages like French \cite{DBLP:conf/acl/LabrakBDRMDG23} and Spanish \cite{carrino-etal-2022-pretrained} have been increasingly made available, German medical data still remains notably underrepresented and has only recently been pushed further \cite{BRESSEM2024121598}.

In this work, several new German biomedical and clinical language models are introduced and extensively evaluated on multiple downstream tasks. All model variants are continuously pre-trained on two different data streams, resulting in public models benefiting from translations of biomedical and medical datasets into German, while private models use internal data from a large German hospital. Translation-based models will be made publicly available.\footnote{\url{https://huggingface.co/ikim-uk-essen}}

\section{Related Work}

Following the success of transformer-based language models, several language models have been developed for biomedical and clinical domains, mainly for English. One of the earliest models is BioBERT \cite{10.1093/bioinformatics/btz682}, which was initialized from a general BERT model and further pre-trained using biomedical data such as PubMed abstracts. This approach demonstrated the effectiveness of continuous pre-training on domain-specific data, allowing the model to more accurately capture the challenging nature of the biomedical domain textual data. Several other specialized models soon followed. One such specialized model is ClinicalBERT \cite{alsentzer-etal-2019-publicly}. While it used a similar continuous pre-training approach to BioBERT, it differed in its integration of clinical data, particularly from sources such as the Medical Information Mart for Intensive Care (\mbox{MIMIC-III},~\citealt{Johnson2016}), a public dataset of de-identified medical records for over 40,000 patients in the intensive care units of Beth Israel Deaconess Medical Center from 2001--2012. Furthermore, \citet{GuTCLULNGP22} introduced PubMedBERT, which was not based on a previously pre-trained model but was trained from scratch on biomedical data. The training dataset consisted of both PubMed abstracts and the full text from PMC, resulting in models that were able to achieve improved performance on a wide range of biomedical tasks. 

In languages other than English, it is more challenging to build such specialized models due to the lack of available data, as is the case for German \cite{zesch_german22}.
However, significant advancements have occurred in this field recently, such as BioGottBERT \cite{Lentzen2022}, a model based on the GottBERT model \cite{DBLP:journals/corr/abs-2012-02110}. GottBERT is a German RoBERTa base model that underwent training utilizing general domain information. BioGottBERT enhanced its medical capabilities by undergoing further pre-training on public German medical texts from Wikipedia and scientific abstracts. This resulted in a significant improvement in performance on medical tasks when compared to GottBERT. In addition to BioGottBERT, the authors trained an ELECTRA \cite{DBLP:conf/iclr/ClarkLLM20} small and basic models from scratch with the aim of evaluating the effectiveness of training new models using only a limited amount of biomedical data. However, the authors reported that this strategy was unsuccessful, and the resulting models were inferior to existing general models. Another German medical model is MedBERTde \cite{BRESSEM2024121598}. This BERT-based model was trained from scratch using various public German medical datasets such as GGPONC 1.0 \cite{borchert-etal-2020-ggponc}, PubMed abstracts, and doctoral dissertations. In addition, the training also incorporated real-world clinical data, such as radiology reports from the Charité University Hospital in Berlin. By integrating such a wide range of clinical and biomedical data, medBERTde aims to provide a comprehensive understanding of the medical field tailored to the German context.

Recently, French has also seen a surge in specialized biomedical and clinical pre-trained language models. Among them is DrBERT \cite{DBLP:conf/acl/LabrakBDRMDG23}, which is based on the RoBERTa architecture and has been trained on both public web data and specialized private medical data from the University Hospital of Nantes. The public web data is a large text corpus called NACHOS (opeN crAwled frenCh Healthcare cOrpuS), crawled from several online biomedical sources. By training on the different datasets, a set of models was obtained, which were then compared by evaluating their performance on a wide range of public and private tasks. Another developed model is AliBERT \cite{berhe-etal-2023-alibert}, a model designed specifically for the French biomedical domain. It was trained using a regularized unigram tokenizer on different sub-corpora of French biomedical textual documents, such as biomedical articles from ScienceDirect, thesis manuscripts, and articles from the Cochrane database. The model excels in F1 and accuracy scores, proving its capabilities in this domain. Interestingly, despite a smaller amount of training data and a shorter pre-training period, AliBERT manages to surpass some notable general French models, highlighting its capabilities.

\section{Pre-Training Datasets}
 
This section details the datasets utilized for pre-training, including clinical data and public medical/biomedical data. While the clinical dataset captures real-world patient insights, the medical data encompasses a wide range of scientific information. These data sources provide the foundation for our models. An in-depth description of these datasets follows.

\subsection{Clinical Data}
The first dataset was sourced from a major German hospital, providing a comprehensive clinical dataset that spans from 2002 to 2023. It includes various clinical documents, including clinical notes, different reports, and doctor's letters. Each text was divided into paragraphs, which were then filtered. Paragraphs with a ratio of letters to other characters below 60\% and paragraphs with an average number of words per line below three were filtered out. The resulting dataset consists of 3,060,845,169 tokens from 25,023,489 documents. To the best of our knowledge, this is the largest German clinical text dataset compiled for pre-training.

\subsection{Public Data}

The second dataset is derived from publicly available biomedical data. It was initiated with approximately 16K German abstracts from PubMed. Recognizing the limited size of this dataset, it was necessary to expand it to improve reliability and coverage. To achieve this, approximately 6 million English PubMed abstracts, along with MIMIC-III clinical notes \cite{Johnson2016}, were translated using the Fairseq WMT'19 English to German translation model\footnote{\url{https://huggingface.co/facebook/wmt19-en-de}, last accessed: 2023-10-13} \cite{ng-etal-2019-facebook}. Although the translation of medical content can be complex and potentially lead to inaccuracies due to specialized terminologies, it provides a way of augmenting the corpus. The decision to use this particular translation model was based on a physician's assessment. They were provided with different translations of 10 PubMed abstract samples and 20 MIMIC notes samples generated by various translation models, including WMT19-en-de, M2M-100 \cite{m2m100}, NLLB \cite{nllb}, T5 \cite{2020t5}, MBart-50 \cite{mbart50}, and OPUS-MT-en-de \cite{opusmt}. Their evaluation guided the final model selection.

Prior to translation, a preprocessing step was performed. All documents were tokenized into individual sentences using the Stanza library \cite{qi-etal-2020-stanza}. These sentences were then grouped into specific segments, each limited to a maximum of 128 tokens. Due to this segmentation, the number of documents increased substantially. For instance, the initial 6 million English PubMed abstracts were divided into approximately 21 million segments or documents for translation. The token limit was chosen based on the observation that segments with more than 128 tokens often suffered from poor translation quality. The number of tokens was determined using the tokenizer of the translation model. This process yielded approximately 45M documents, detailed in Table \ref{table:public_dataset}.

\begin{table}[ht]
    \centering
    \label{table:public_dataset}
    \begin{tabular}{lrr}
        \toprule
        \textbf{Dataset} & \textbf{Tokens} & \textbf{Documents} \\
        \midrule
        German PubMed & 5M  & 16K \\ 
        PubMed & 1,700M & 21M \\
        MIMIC-III & 695M & 24M\\
        \midrule
        Total & 2,400M & 45M\\
        \bottomrule
    \end{tabular}
    \caption{Public dataset composition. The increase in the number of documents for PubMed and MIMIC-III compared to the original source is due to the segmentation of the content into chunks of 128 tokens or less for the translation process.}
\end{table}

\begin{figure*}[ht]
    \centering
    \includegraphics[width=1\linewidth]{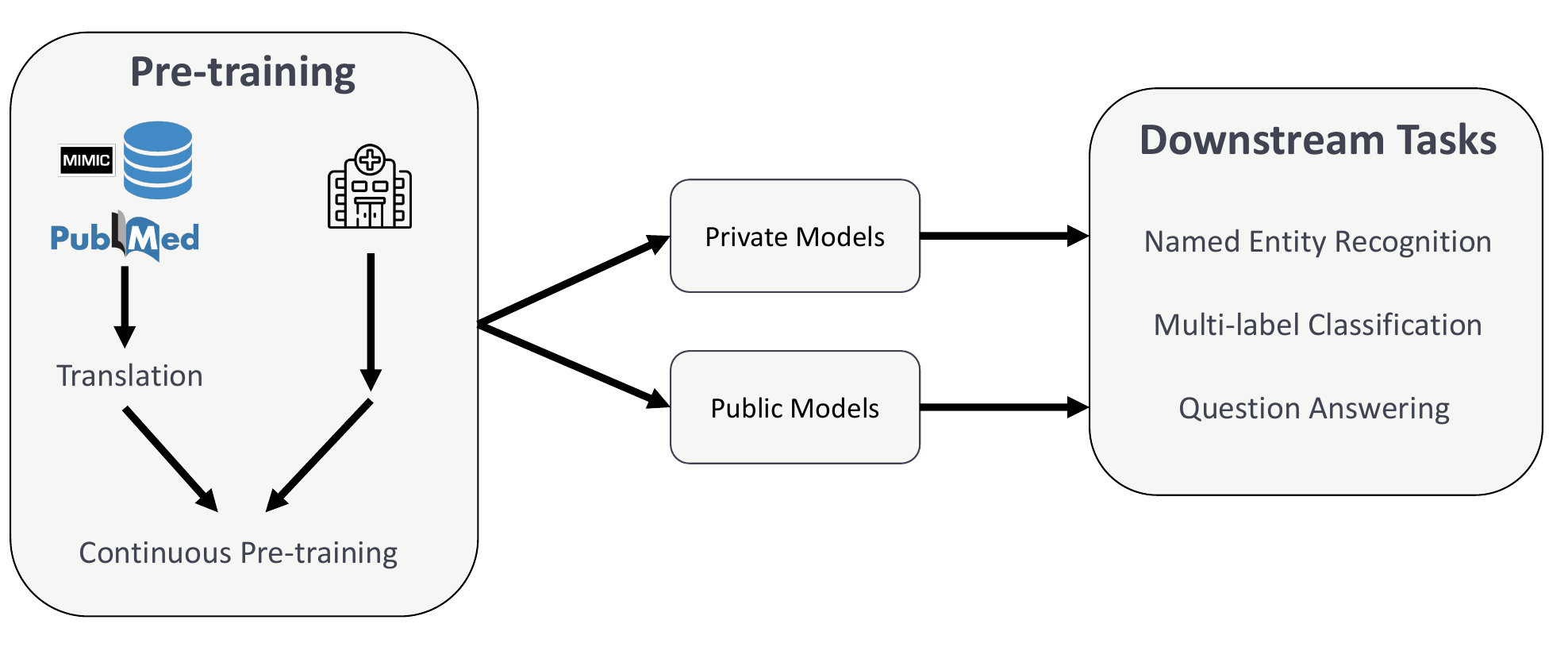}
    \caption{Workflow for Continuous Pre-training of different Publicly Available General Domain Models with Private and Public Datasets: The private dataset comes from a major German hospital and yields 25M documents. The public dataset, sourced from English PubMed abstracts and MIMIC-III clinical notes, is preprocessed, tokenized into sentence segments, and translated into German via the Fairseq WMT'19 English to German model, resulting in approximately 45M documents. Both model setups are subject to model fine-tuning and evaluation across various biomedical and clinical German language downstream tasks.}
    \label{fig:overview-label}
\end{figure*}

\section{Base Models}

In this section, the models included in our benchmark are described. As a baseline for our evaluations, we use models that have been pre-trained on extensive datasets and have demonstrated strong performance on a variety of general and medical NLP tasks. The general German language models GBERT$_{base}$, GBERT$_{large}$ and GELECTRA$_{large}$ \cite{chan-etal-2020-germans} were primarily considered. They were trained on four different datasets: the German portion of the \textit{Open Super-large Crawled ALMAnaCH coRpus} (OSCAR) \citep{ortiz-suarez-etal-2020-monolingual}, German Wikipedia dump, \textit{The Open Parallel Corpus} (OPUS) \citep{tiedemann-2012-parallel}, and Open Legal Data \citep{10.1145/3383583.3398616}. Another model is GottBERT \citep{DBLP:journals/corr/abs-2012-02110}, a German RoBERTa$_{base}$ model trained on the German part of OSCAR data. Furthermore, two multilingual models XLM-R \cite{xlmroberta} and mDeBERTa V3 \cite{debertav3} were also considered. Additionally, GeBERTa$_{base}$ and GeBERTa$_{large}$ \cite{dada2023impact} were explored, with these DeBERTa v2 \cite{DBLP:conf/iclr/HeLGC21} based models pre-trained from scratch on a combination of data sources ranging from Wikipedia to medical datasets. Finally, the German biomedical and clinical models BioGottBERT \cite{Lentzen2022} and medBERTde \cite{BRESSEM2024121598} were also examined to provide additional comparative insights.

\section{Pre-training}

Continuous pre-training of several publicly available general domain models was performed with the aforementioned datasets.
First, we continued the pre-training of the GeBERTa$_{base}$ and GeBERTa$_{large}$ models with the clinical dataset. The objective was to quantify the contribution of clinical data to the performance of the models. While the original model has already been trained on medical texts obtained mainly from MIMIC-III, translated PubMed abstracts, and filtered CC100 \cite{wenzek-etal-2020-ccnet}, it has not seen any real-world German clinical data. Both models were trained for 200k AdamW \cite{adamw} optimization steps with a batch size of 512. The learning rate was set to 3e$^{-5}$ for the base model and 2e$^{-5}$ for the large model. Additionally, the GBERT$_{base}$ and GBERT$_{large}$ models were further pre-trained on the clinical data with the same parameters. These models were not explicitly pre-trained on medical texts before.

In order to separately quantify the influence of translation-based medical texts and clinical documents, additional pre-training experiments were conducted with the GBERT$_{base}$ and GBERT$_{large}$ models. Both models underwent further pre-training on the translated dataset. The GBERT${large}$ was trained for 73k steps with a learning rate of 5e$^{-5}$ and a batch size of 144. On the other hand, the GBERT${base}$ was trained for 75k steps using a batch size of 336. Due to hardware limitations, the initial experiments were conducted separately, reflecting the differences in batch sizes between the experiments. The AdamW optimizer was also employed during this pre-training for the optimization process.

\section{Downstream Datasets \& Tasks}

To evaluate the pre-trained models, the models were fine-tuned and subsequently evaluated across a range of downstream tasks, aiming to determine their efficacy and adaptability in specialized biomedical and clinical domains.

 \subsubsection*{BRONCO}
The Berlin-Tübingen-Oncology Corpus (BRONCO) \cite{10.1093/jamiaopen/ooab025} is a comprehensive, freely accessible German corpus derived from 200 oncology discharge summaries of cancer patients. These summaries have been manually de-identified and annotated to highlight key entities such as diagnoses, treatments, and medications. This annotated corpus contains 11,434 sentences and 89,942 tokens, with 11,124 annotations identifying medical entities suitable for named entity recognition (NER). While the authors have released 75\% (or 150 summaries) of the dataset to the public, they have kept 25\% (or 50 summaries) as a held-out set to ensure unbiased evaluation or data contamination. Given the unavailability of the BRONCO50 dataset, a 5-fold cross-validation was performed to train and evaluate models on the BRONCO150 dataset.
 
 %ggponc
 \subsubsection*{GGPONC 2.0}
 The German Guideline Program in Oncology NLP Corpus (GGPONC) 2.0 \cite{borchert-etal-2022-ggponc} represents a significant advance in German medical language resources and offers a large corpus for NER applications. Based on the top-level hierarchies of the SNOMED CT concept model, its annotation scheme distinguishes between several subclasses of entities. The ``Finding'' category includes entities such as diagnosis, pathology, and other relevant findings. The ``Substance'' category delves into clinical drugs, nutrients or body substances, and external substances. In addition, the ``Procedure'' category houses entities associated with therapeutic and diagnostic procedures. Recognizing the complex nature of clinical texts, where entity boundaries are often ambiguous, GGPONC 2.0 is complemented by a comprehensive annotation guide that clarifies the definition of each entity class.

 %grascco
 \subsubsection*{GraSCCo}
Graz Synthetic Clinical text Corpus (GraSCCo) \cite{Modersohn2022} is a synthetic German corpus consisting of about 60 clinical documents with more than 43,000 tokens. It includes a series of alienation steps to hide privacy-sensitive information in real clinical documents, the true origin of all GraSCCo texts. As a result, the data is publicly available without any legal restrictions. Within the medbert.de paper, an additional annotation of the GraSCCo data for an NER task was performed by the authors of the GGPONC 2.0. We use these annotations for our benchmark as well.
 
 %clef
 \subsubsection*{CLEF eHealth 2019}
The CLEF (Conference and Labs of the Evaluation Forum) \cite{Kelly2019} eHealth dataset is a curated collection of non-technical summaries (NTS) of animal experiments from Germany, which was used to organize the Multilingual Information Extraction Task (Task 1) in the CLEF eHealth Challenge 2019 \cite{dorendahl2019overview}. These NTS have been made publicly available to increase transparency in animal research. For the identification of the primary diseases that are the focus of the experiments, each NTS in the dataset is manually annotated with the corresponding ICD-10 codes. Reports are predominantly scientific and biomedical, while some clinical jargon could also be observed.  In total, the dataset contains over 8,000 NTSs for training and an additional 407 NTSs for testing. The primary objective associated with this dataset is a multi-label classification, where systems are challenged to assign the relevant ICD-10 codes to each summary automatically.

 %UK essen QnA
\subsubsection*{RadQA}
The RadQA dataset \cite{dada2024information} is an extractive question-answering dataset created from 1,223 anonymized radiology reports of brain CT scans from a large hospital in Germany. For its development, three medical student assistants in their sixth and eighth semesters were assigned to annotate 29,273 question-answer pairs. The annotators were provided with a list of questions designed with the input of a radiologist. This decision was influenced by the unique nature of radiology queries and the challenges of annotating sensitive clinical data. The annotators generated one custom question for every third report to ensure variety.

\section{Fine-tuning}
In a comprehensive evaluation, the performance of the models was assessed on a variety of downstream tasks. The problem classes included NER (BRONCO, GGPONC 2.0, GraSCCO), multi-label classification (CLEF eHealth 2019), and extractive question answering (RadQA).

\subsubsection*{Hyperparameters}
Choosing the right hyperparameters is crucial for optimizing model performance. However, in this study, an extensive hyperparameter search was intentionally avoided to reflect a clinical environment with limited computational resources. Instead, a mixture of standard and task-specific settings was opted for, starting from the default parameters in HuggingFace. Only the learning rate and batch size were adjusted to address training instabilities, and the number of epochs was set to ensure convergence of all models, resulting in variation across different dataset sizes in the benchmark.

While this approach may not yield the optimal results that can be achieved by an extensive grid search with hundreds of configurations, it provides valuable insights into the performance of models under standard parameter conditions, which is particularly relevant for clinical applications where computational resources are often limited. By using uniform configurations for basic and large models and an appropriate small number of epochs for fine-tuning, comparability across the models and tasks in the benchmark is aimed to be increased.

Although a more in-depth analysis of the hyperparameter optimization process could further enhance the rigor of this work, the methodology used reflects a practical scenario in clinical settings and offers a realistic assessment of model performance under resource constraints. The specific hyperparameters used for each downstream task are listed in Table \ref{tab:hyperparameters}.

\begin{table}[t]
\centering
\resizebox{\columnwidth}{!}{
\begin{tabular}{lccc}
\toprule
    \multirow{2}{*}{Task} & \multicolumn{1}{c}{Learning Rate} & \multicolumn{1}{c}{Batch Size} & \multirow{2}{*}{Epochs} \\
\cmidrule(lr){2-2} \cmidrule(lr){3-3}
& base, large & base, large & \\
\midrule
BRONCO              & $3e-5, 1e-5$ &  $16, 16$ & $20$ \\
GGPONC 2.0          & $3e-5, 1e-5$ &  $16, 16$ & $5$ \\
GraSCCo             & $3e-5, 1e-5$ &  $16, 16$ & $20$ \\
CLEF eHealth   & $4e-5, 1e-5$ &  $16, 32$ & $20$ \\
RadQA                 & $3e-5, 1e-5$ &  $16, 16$ & $10$ \\
\bottomrule
\end{tabular}
}
\caption{Hyperparameters of the different downstream tasks. 
%$\dagger$ denotes default parameter \tz{a bit small}
}
\label{tab:hyperparameters}
\end{table}

\subsubsection*{Task Specifics}
In the context of the CLEF eHealth 2019 fine-tuning, class imbalances among the labels are addressed with logarithmic weighting. In multi-label scenarios, there are often many classes, among which some appear rarely and others very frequently. This can lead to training not converging at all or being sensitive to hyperparameters. Some model configurations were unstable for this task, especially for large models. Using the adjusted weights led to stable results. The positive class weights, \(w_j\), are calculated for label \(j\) and use the following logarithmic scheme:

\[ w_j = \log\left( \frac{N}{1 + c_j} \right) \]

Where \(N\) denotes the total number of training samples and $c_j$ is the count of each label. The logarithmic weighting scheme adjusts weights inversely to the label frequency in the training data, ensuring balanced attention between frequent and rare labels during training.

\section{Results}
The fine-tuning results for all downstream tasks are split between Table \ref{pref1} and Table \ref{table:models_performance}. 

Across all tasks, our clinically pre-trained models achieved the highest F1-Scores, with the exception of GGPONC 2.0, where one of the translation models achieved the highest score. This is especially evident in the results of the GraSCCo dataset, where clinical pre-training improved the performance of GBERT$_{base}$ by 5.4 percentage points in F1-Score and the performance of GBERT$_{large}$ by 6.9 percentage points. In addition, the GBERT-BioM-translation models were able to outperform the general models. For instance, pre-training on translated texts resulted in an improvement of 2.1 percentage points over GBERT$_{base}$. 

Although BRONCO150 is the only NER task with real-world clinical documents, there is no observable performance difference between MedBERTde, GeBERTa$_{base}$, and our clinical base models. However, the GeBERTa-Clinical$_{large}$ model outperformed the second-best model by 0.5 percentage points. The minor differences between the various models and the lack of clear advantages of clinical models can be attributed to the small size of the dataset. Interestingly, on GGPONC 2.0, various general domain models are on par with clinical models or even better. In the case of GeBERTa$_{base}$, the additional clinical pre-training decreased its performance. It is worth noting that the dataset consists of guidelines for oncologists written in a different writing style than clinical documents. Overall, the translation-based models achieved the highest results.

In the multi-label classification CLEF eHealth 2019 task, large models generally performed better. For example, GBERT$_{large}$ has an F1-Score that is around 1.6 percentage points better than its base variant. Domain-specific continuous pre-training on the translations reached strong performance across all metrics. Prioritizing a balance between precision and recall favors the GELECTRA$_{large}$ model, while GeBERTa-Clinical$_{large}$ surpasses in precision. The general domain GBERT$_{large}$ model provides the highest recall, although its precision is lower compared to others.

Looking at the difference between GBERT-Clinical$_{base}$ and GBERT-BioM-Translation$_{base}$ in Table \ref{table:models_performance}, the latter achieves comparable results to its clinical counterpart on BRONCO, GGPONC 2.0, and RadQA. Only on GraSCCo the results of the GBERT models trained on the translations are worse. Compared to the baseline GBERT general models, all further pre-trained models have an advantage. In this case, we see no clear advantage of training on German clinical data over translated texts.

In summary, across almost all tasks, clinical pre-training and translation-based pre-training led to better performance than general domain models that were not explicitly trained on medical data. While the results indicate a slight advantage for models trained on our clinical data, the performance difference tends to be small, and in some cases, the translation-based models even outperform the clinical ones.

\begin{table*}[t]
\centering
%\resizebox{\textwidth}{!}{%
\begin{tabular}{llccccccccc}
    \toprule
    & \multirow{2}{*}{Model} & \multicolumn{3}{c}{CLEF eHealth 2019} & \multicolumn{2}{c}{RadQA} & \multicolumn{3}{c}{GraSCCo} \\
    \cmidrule(lr){3-5} \cmidrule(lr){6-7} \cmidrule(lr){8-10}
    & & F1 & P & R & F1 & EM  & F1 & P & R \\
    \midrule
    & GBERT$_{base}$ \cite{chan-etal-2020-germans}                   & .816 & .818 & .815 & .794 & .707  & .642 & .617 & .676 \\
    & GBERT$_{large}$ \cite{chan-etal-2020-germans}                     & .832 & .802 & \textbf{.865} & .809 & .718  & .647 & .617 & .680 \\
    & GottBERT  \cite{DBLP:journals/corr/abs-2012-02110}                 & .791 & .818 & .765 & .796 & .712  & .652 & .681 & .624 \\
    & XLM-R$_{large}$ \cite{xlmroberta}              & .804 & .781 & .829 & .813 & .731  & .674 & .655 & .694 \\
   % gelectra-base                   & .746 & .730 & .763 & - & - & - & - & -  \\
    & GELECTRA$_{large}$  \cite{chan-etal-2020-germans}               & .827 & .826 & .828 & .812 & .725 & .681 & \textbf{.702} & .661  \\
    \multirow{-6}{*}{\rotatebox{90}{general}} & mDeBERTa V3$_{base}$ \cite{debertav3}               & .793 & .786 & .801 & .810 & .741 & .675 & .646 & .706 \\ 
    \midrule
    & GeBERTa$_{base}$  \cite{dada2023impact}                   & .823 & .817 & .829 & .839 & .769 & .684 & \textbf{.702} & .667 \\
    & GeBERTa$_{large}$  \cite{dada2023impact}                  & .837 & .848 & .826 & .834 & .757 & .669 & .700 & .640  \\ 
    & BioGottBERT  \cite{Lentzen2022}               & .791 & .779 & .803 & .797 & .706 & .637 & .673 & .605  \\
    &  GBERT-BioM-Translation$_{base}$$^{\dagger}$        & .825 & .851 & .801 & .808 & .716 & .661 & .642 & .681  \\ 
    \multirow{-4}{*}{\rotatebox{90}{medical}} & GBERT-BioM-Translation$_{large}$$^{\dagger}$         & .833 & .860 & .807 & .811 & .714 & .692 & .677 & .707  \\ 
    % & de-longformer-translation$^{\dagger}$       & .833 & .815 & .852 & - & - & .622 & .598 & .648 \\
    \midrule
    & MedBERTde \cite{BRESSEM2024121598}         & .836 & .839 & .833 & .833 & .761 & .660 & .626  & .697 \\
    & GBERT-Clinical$_{base}$$^{\dagger}$           & .833 & .853 & .815 & .807 & .726 & .696 & .670 & .725 \\
    & GBERT-Clinical$_{large}$$^{\dagger}$          & \textbf{.843} & \textbf{.876} & .812 & .806 & .710 & \textbf{.716} & .692 & \textbf{.742} \\
    & GeBERTa-Clinical$_{base}$$^{\dagger}$           & .804 & .816 & .792 & .845 & .762 & .680 & .699 & .662  \\
    
    % gbert-base-translated           & .837 & .836 & .838 & .803 & .706 & - & - & -  \\ 
    % gbert-large-translated          & \textbf{.838} & .835 & .842 & .809 & .716 & - & - & - \\ 
    \multirow{-5}{*}{\rotatebox{90}{clinical}}& GeBERTa-Clinical$_{large}$$^{\dagger}$          & .833 & .874 & .796 & \textbf{.846} & \textbf{.768} & .703 & .677  & .730 \\
    \bottomrule
\end{tabular}
%}
\caption{Performance of Different Models on various Downstream Tasks: CLEF eHealth 2019 (Multi-label classification), RadQA (Extractive Questions Answering), and GraSCCo (NER). F1-Scores reported are micro-averaged. P denotes Precision, R denotes Recall, EM denotes Exact Match, $\dagger$ marks models that we pre-trained}
\label{pref1}

\end{table*}

\begin{table*}[t]
\centering
\begin{tabular}{llccc ccc}
    \toprule
    &\multirow{2}{*}{Model} & \multicolumn{3}{c}{BRONCO150} & \multicolumn{3}{c}{GGPONC 2.0} \\
    \cmidrule(lr){3-5} \cmidrule(lr){6-8}
    & & F1 & P & R & F1 & P & R  \\
    \midrule
    & GBERT$_{base}$                     & .833 $\pm$ .004 & .818 $\pm$ .002 & .849 $\pm$ .011 & .770 & .761 & .780  \\
    & GBERT$_{large}$                    & .835 $\pm$ .006 & .820 $\pm$ .004 & .852 $\pm$ .011 & .772 & .758 & .786 \\
    & GottBERT                   & .840 $\pm$ .008 & .827 $\pm$ .010 & .854 $\pm$ .012 & .756 & .744 & .768  \\
    & XLM-R$_{large}$               & .841 $\pm$ .003 & .823 $\pm$ .007 & .860 $\pm$ .007 & .775 & .764 & .786  \\
    & GELECTRA$_{large}$                  & .850 $\pm$ .006 & \textbf{.835 $\pm$ .005} & .865 $\pm$ .007 & .780 & .769 & .792  \\
    \multirow{-6}{*}{\rotatebox{90}{general}} & mDeBERTa V3$_{base}$               & .843 $\pm$ .005 & .824 $\pm$ .007 & .862 $\pm$ .007 & .768 & .753 & .783  \\ 
    \midrule
    & GeBERTa$_{base}$                    & .848 $\pm$ .007 & .830 $\pm$ .010 & .866 $\pm$ .007 & .772 & .761 & .783  \\
       & GeBERTa$_{large}$                    & .847 $\pm$ .008 & .825 $\pm$ .003 & .872 $\pm$ .001 & .772 & .758 & .786  \\
    & BioGottBERT              & .844 $\pm$ .011 & .826 $\pm$ .012 & .863 $\pm$ .013 & .770 & .756 & .785  \\
     & GBERT-BioM-Translation$_{base}$$^{\dagger}$           & .842 $\pm$ .006 & .824 $\pm$ .007 & .861 $\pm$ .007 & .780 & .766 & \textbf{.794}   \\ 
   \multirow{-4}{*}{\rotatebox{90}{medical}} & GBERT-BioM-Translation$_{large}$$^{\dagger}$          & .844  $\pm$ .007 & .825 $\pm$ .007 & .864 $\pm$ .009 & \textbf{.786} & \textbf{.779} & .793  \\ 
    %\multirow{-5.5}{*}{\rotatebox{90}{medical}} & de-longformer-translation$^{\dagger}$       & - & - & - & - & - & - \\
    \midrule
    & MedBERTde          & .848 $\pm$ .005 & .833 $\pm$ .008 & .864 $\pm$ .006 & .755 & .744 & .744 \\
    & GBERT-Clinical$_{base}$$^{\dagger}$           & .847 $\pm$ .009 & .828 $\pm$ .009 & .867 $\pm$ .011 & .772 & .763 & .781  \\
    & GBERT-Clinical$_{large}$$^{\dagger}$          & .844 $\pm$ .007 & .825 $\pm$ .007 & .864 $\pm$ .009 & .785 & .778 & .792  \\
    & GeBERTa-Clinical$_{base}$$^{\dagger}$           & .846 $\pm$ .006 & .823 $\pm$ .007 & .870 $\pm$ .009 & .768 & .757 & .780 \\
    % gbert-base-translated           & .836 & .819 & .853 & .775 & .764 & .787 & .653 & .676 & .632 \\
    % gbert-large-translated          & .847 & .826 & .869 & \textbf{.788} & \textbf{.774} & \textbf{.801} & .666 & .644 & .689 \\
    \multirow{-5}{*}{\rotatebox{90}{clinical}} &  GeBERTa-Clinical$_{large}$$^{\dagger}$          & \textbf{.855 $\pm$ .007} & .832 $\pm$ .006 & \textbf{.878 $\pm$ .011} & .773 & .760 & .787 \\
    \bottomrule
\end{tabular}
\caption{Performance of Different Models on NER Downstream Tasks: BRONCO150 (5-fold cross validation), GGPONC 2.0. F1-Scores reported are micro-averaged. P denotes Precision, R denotes Recall, $\dagger$ marks models that we pre-trained}
\label{table:models_performance}

\end{table*}

\section{Discussion and Limitations}
Overall, the results show that especially the addition of medical pre-training data gives a performance advantage, but not necessarily the quality of the data. For example, a 6.9 percentage points improvement in F1-Score performance was measured on the GraSCCo dataset between GBERT$_{large}$ and GBERT-Clinical$_{large}$. However, the difference between GBERT-Clinical$_{large}$ and GBERT-BioM-Translation$_{large}$ is only 2.4 percentage points. Similarly, in RadQA the difference between GBERT$_{base}$ and GBERT-BioM-Translation$_{base}$ is 1.4 percentage points, but the difference between the two further pre-trained models is only 0.1 percentage points. We conclude that the presence of medical data is crucial for pre-training, but not necessarily its quality or proximity for the downstream task. The small differences between our GeBERTa-Clinical and standard GeBERTa models further support this hypothesis, as the standard version of GeBERTA already contained translations. In this context, it is also worth noting that the difference between GBERT-Clinical and GBERT-BioM-Translation models is particularly evident on the GraSCCo task, even though it consists of synthetically generated documents and not real clinical texts. In contrast, smaller differences are evident in clinical tasks such as RadQA.

These findings open up possibilities to train clinically applicable models in different scenarios where only limited clinical data is available or cannot be accessed for privacy reasons. This is the case, for example, with low-resource languages, where far fewer clinical documents are written. Additionally, synthetic or translated public data can also help protect patient privacy by avoiding pre-training on patient data.

Despite being less resource-intensive than training from scratch, this work was able to achieve good results on various downstream tasks. This highlights the effectiveness of transfer learning and the value of pre-trained models.

As we discuss the findings and methodologies, it is important to address some of the challenges and limitations. Determining the optimal hyperparameters can be intricate. Subtle changes, such as adjusting the batch size or seed, can affect the results. As a result, direct comparisons with previous work should be made with caution, especially when score variances are minimal. Additionally, although models grounded on translations can be made public, sharing models trained on proprietary clinical data remains prohibited. This restriction is grounded in data protection measures and concerns about potential retrieval attacks that might expose the training data.

\section{Conclusion}
In this study, several new German biomedical and clinical language models were introduced coming from two data streams: public translation data and private clinical data from a large German hospital, with the translation-based models made publicly available to the research community. These models were assessed on five downstream tasks and compared to a variety of German and multilingual models from the general, biomedical, and clinical domains.
It was demonstrated that the translation of PubMed is a promising approach that can avoid data protection concerns. In most downstream tasks, translation-based models achieved comparable results to large-scale clinical models, although only 6 million translated abstracts were utilized. In particular, clinical downstream tasks show that domain-specific pre-training can still be worthwhile, but the general domain German models, such as GBERT, can be sufficient in many domains. Despite the performance and ease of distribution for translation-based models, it is important to recognize that in half of the tasks tested, models derived from private clinical data still performed better, highlighting the importance and effectiveness of large specialized data sources. 

Future work could look at training models that use all available PubMed abstracts. In addition, we aim to explore the training of German medical large language models and leverage their capabilities.

\section{Ethical Statement}
This study was conducted in alignment with the principles of the Helsinki Declaration and received approval from the Institutional Review Board (IRB). Throughout the research, the team maintained transparency, integrity, and respect for the unique requirements of medical data, emphasizing patient welfare and data protection standards.

The use of language models in the medical field raises several ethical concerns. Biases in the training data can lead to poor outcomes for underrepresented groups, posing a significant issue in healthcare delivery. This highlights the need for strategies to identify and address potential biases, ensuring equitable representation and outcomes for all patient groups.

Moreover, the ``black box'' nature of these models complicates their use in medical decision-making, where transparency and trust are crucial. This emphasizes the importance of prioritizing interpretability and explainability in the development and application of language models, fostering trust and enabling informed decision-making.

Furthermore, the use of patient data presents challenges in informed consent and privacy, especially given the difficulties in securing individual consent for large datasets and the inability for patients to opt-out post-training. This underscores the necessity for robust data protection measures, adherence to relevant privacy regulations, and ensuring that patient data is handled with the utmost care and confidentiality.

These issues emphasize the need for careful ethical considerations in the deployment of language models in healthcare, and this study aimed to address these concerns through a comprehensive and ethically grounded approach.

\section{Acknowledgement}

The work of Ahmad Idrissi-Yaghir and Henning Schäfer was funded by a PhD grant from the DFG Research Training Group 2535 Knowledge- and data-based personalisation of medicine at the point of care (WisPerMed).

\section{Bibliographical References}\label{sec:reference}

\bibliographystyle{lrec-coling2024-natbib}
\bibliography{lrec-coling2024-example}

% \section{Language Resource References}
% \label{lr:ref}
% \bibliographystylelanguageresource{lrec-coling2024-natbib}
% \bibliographylanguageresource{languageresource}

\end{document}